# Ear Identification by Fusion of Segmented Slice Regions using Invariant Features: An Experimental Manifold with Dual Fusion Approach

Dakshina Ranjan Kisku*[a], Phalguni Gupta[b], Jamuna Kanta Sing[c]
[*a]Dr. B. C. Roy Engineering College, Durgapur – 713206, West Bengal, India
[b]Indian Institute of Technology Kanpur, Kanpur – 208016, Uttar Pradesh, India
[c]Jadavpur University, Kolkata – 700032, West Bengal, India

**ABSTRACT**

This paper proposes a robust ear identification system which is developed by fusing SIFT features of color segmented slice regions of an ear. The proposed ear identification method makes use of Gaussian mixture model (GMM) to build ear model with mixture of Gaussian using vector quantization algorithm and K-L divergence is applied to the GMM framework for recording the color similarity in the specified ranges by comparing color similarity between a pair of reference ear and probe ear. SIFT features are then detected and extracted from each color slice region as a part of invariant feature extraction. The extracted keypoints are then fused separately by the two fusion approaches, namely concatenation and the Dempster-Shafer theory. Finally, the fusion approaches generate two independent augmented feature vectors which are used for identification of individuals separately. The proposed identification technique is tested on IIT Kanpur ear database of 400 individuals and is found to achieve 98.25% accuracy for identification while top 5 matched criteria is set for each subject.

## 1. INTRODUCTION

Biometric technologies [1] can be used for identity verification of users by their physiological or behavioral characteristics. Biometric systems are divided into two modalities: monomodal [1] and multimodal biometric [2] systems. Monomodal biometric system [1] uses biometric cue of single biometric characteristic while multimodal biometric system uses two or more biometric cues in an integrated form. Many monomodal biometric systems have been proposed that include fingerprint, iris, face, palmprint and hand geometry, signature, ear, retina, DNA etc and on the other hand, there are many multibiometric systems [2] such as face and fingerprint [3], face and iris [3], face, fingerprint and hand geometry [3], voice and signature [3], etc. However, irrespective of well accepted and traditional biometric systems, ear biometric [4,5,6] is considered as a new biometric trait. Ear biometric is often compared with face biometric [7] trait. Unlike face biometrics with changing lighting conditions, non-uniform distributions of intensity and spatial resolution according to illumination, variability with various expressions and different head positions, ear biometrics have several advantages over facial features. The ear shape [5,6] does not change over time and ageing. Moreover less effect of lighting conditions and spatial distributions of pixels have been made ear biometrics an emerging area in biometric identity verification.

Any biometric system [1] is composed of methods of feature extraction and representation, classification, and finally recognition. Without using distinct and invariant features for representation of biometric template higher accuracy can not be achieved. Therefore, the invariant features, which are extracted from higher matching probability texture regions of biometrics template [1], can be one of the robust methodologies for development of strong biometrics system.

Identifying or verifying people using ear recognition has been increasing in the last few years. Therefore, the ear biometrics can be used as a robust recognition model when invariant features are used. There are many ear recognition systems proposed in the literatures. Voronoi diagram with curve segments are used in [8] for automatic ear recognition. An interesting ear recognition system has been proposed in [9] where force field transformation is used. Authors of [13] have proposed a novel approach in 3D ear recognition system using local surface descriptor. A comparative study has been made in [10] for ear and face biometrics.

In initial phase of object recognition, many pattern classification techniques with various feature representation approaches have been proposed in [11,12]. Due to thrust for robust object recognition, many researchers have still faith on invariant features. As a result, some exciting 2D and 3D ear recognition techniques [5,6,13] have been proposed by using the general object recognition techniques [11,12]. In these techniques [5,6,13] ear recognition techniques have been developed under varying lighting conditions and poses with varying performances. Performances often are degraded due to uncontrolled lighting conditions and changeable camera viewpoints. Some widely used pattern representation techniques like PCA based [10] and LDA-kernel based [6] ear recognitions have been considered. In the former technique, Principal Component Analysis has been applied to ear recognition, where the recognition performance is achieved under limited conditions while the later technique uses the combination of Linear Discriminant Analysis (LDA) and kernel techniques which overcomes the limitations posed by PCA. However, the problem of generalized eigenvalues is not being solved in the later work [6]. Some feature based and geometric measurement based techniques have been applied to ear recognition successfully in [8,9]. In [14], authors have applied block-based multi-resolution techniques for ear recognition using wavelet transform and Local Binary Pattern (LBP). Unfortunately, these techniques are very sensitive to occlusion and misregistration. Moreover, occlusion in particular often occurs due to the ear obscured by hair or earrings. To deal with these problems, some invariant SIFT features based techniques have been proposed. In [5], the authors have presented a robust ear recognition system that deals with not only background clutter and occlusion, but also with pose variations. In this work, ear is considered as a planer surface and creating a homography transform using SIFT features which lead to be ears being made registered accurately. The matching of SIFT features reduces the gallery size and enable a precise ranking. Another SIFT feature based ear recognition proposed in [15] where 16 keypoints for each ear image are generated and matching is done between a pair of ear images by calculating the number of keypoint matches using closest square distance. A feature level fusion approach has been applied to ear recognition [16]. In this work, SIFT features are extracted from the ear images of different poses and finally these SIFT features are fused and matching is performed using Euclidean distance.

This paper presents an invariant SIFT descriptor [12] based ear identification and verification system that is useful for pose variations and illuminations. However, the IIT Kanpur ear database [17] consists of ear images of almost pose invariant and little illumination variant. Therefore, the effect of the first factor can be negligible and the present system deals with illumination problem. In addition to this, it has been determined that while the features are detected from clustered regions where intensity variations are minimum, the detected features are more useful for identification and verification work rather than the features which are detected from the whole subject's pattern without color homogeneity consideration in particular. Image clustering is useful in image analysis and pattern recognition. Image segmentation partitions an image into non overlapping regions. A region is defined as a homogeneous group of connected pixels with respect to a chosen property in terms of color, gray levels, motion, texture, etc. Among them, color image clustering attracts more attention because it can provide more rich spatial information than other properties. However, the proposed system uses gray scale clustered ear image which can be found by transforming the clustered color ear image. SIFT descriptor is used to extract the invariant keypoints from the clustered ear image.

The proposed method uses Gaussian mixture model [18] for modeling the skin color of the ear image and uses K-L divergence algorithm [19] to cluster the whole ear image into a number of color slice regions by recording the color similarity properties from a pair of ear images. From the clustered ear image, SIFT keypoint features [12] are extracted from each grayscale slice region. To improve the robustness and performance of the system two feature fusion techniques such as concatenation and Dempster-Shafer decision theory are used to fuse the invariant features extracted from the slice regions. Finally, authenticity has been established by using two distance measures namely Euclidean distance [17] and Nearest Neighbor approaches [17]. The proposed system has been evaluated with the IIT Kanpur ear database.

The paper is organized as follows. Section 2 discusses ear image modeling using Gaussian Mixture Model while K-L divergence used to record the color similarity regions is presented in Section 3. Section 4 presents SIFT features extraction process from each gray slice region. Fusion strategies of keypoint features using concatenation and Dempster-Shafer decision theory is presented in Section 5. Section 6 presents the experimental results of the proposed identification and verification. Conclusion is given in the last section.

## 2. EAR MODELING USING GAUSSIAN MIXTURE MODEL

To model an ear image, Gaussian Mixture Model [18] is used. An ear image is considered as a collection of coherent regions. Each homogeneous color region is represented by a Gaussian distribution in the image plane and Gaussian mixture model refers to the set of all color slice regions. Therefore, an ear can be a mixture of Gaussian models and mixture model deals with color features in the color feature space, in particular. For segmentation of color features in the feature space in terms of pixels in detected ear image based on the probabilities of identical color spaces, vector quantization is applied to cluster the color features of pixels. Vector quantization [20] can be considered as a fitting model where the clusters are represented by conditional density functions. In this fitting model, predetermined set of probabilities are the weights. Data contained within vector quantization framework can be fitted with Gaussian mixture model and the probability density function of a dataset is represented as a collection of Gaussians. This convention can be represented by the following equation:

$$f(x) = \sum_{i=1}^{N} P_i f(x \mid i) \quad (1)$$

where $N$ is the number of clusters or slice regions in ear image, $p_i$ is the prior probability of cluster $i$ and $f_i(x)$ is the probability density function of cluster $i$. The conditional probability density function $f_i(x)$ can be represented as

$$f(x \mid i) = \frac{\exp(-\frac{1}{2}(x - m_i)^t \sum_i^{-1} (x - m_i))}{(2\pi)^{P/2} \mid \sum_i \mid^{1/2}} \quad (2)$$

where $x \in R^P$, and $m_i$ and $\sum_i$ are the mean and covariance matrix of cluster $i$, respectively. To determine the maximum likelihood parameters of a mixture of $i$ Gaussians the Expectation-Maximization (EM) algorithm [18] is used and the Minimum Description Length (MDL) principle is used further to select the values of $i$ while $i$ ranges from 3 to 6.

## 3. KULLBACK-LEIBLER DIVERGENCE FOR COLOR SIMILARITY MEASUREMENT

Kullback-Leibler (K-L) divergence [19] is given as non-symmetric distance measure between probability distributions. In computer vision and pattern classification, it is often needed to compute the similarity between two images or coherent regions of two images. It is performed by matching the spatial features or color features of the images. K-L divergence measures the theoretic criterion that gives a dissimilarity score between the probabilities densities of two images or regions of images. It measures the expected number of extra bits required to code samples from one probability distribution when using a code based on another probability distribution, rather than using a code based on the first distribution. Therefore, the first distribution model represents the "true" distribution of data, observations, or a precise calculated theoretical distribution. The second probability distribution measure typically represents a theory, model, description, or approximation of the first one.

Assume a representation of an ear image is given by Gaussian mixture model density function and to define a similarity measure between two ear images in terms of the corresponding two Gaussian density models, Kullback-Liebler (K-L) divergence measure is used. Once Gaussian mixture models [18] for color pixels have been formed in the cropped ear images K-L divergence is used for keep color consistency in the coherent color slice regions independently and is also used for finding similarity among the ear images in terms of mixture of Gaussian models.

The K-L divergence can be defined between two probability density functions $p(x)$ and $q(x)$ found from two color ear images,

$$KL(p \parallel q) \stackrel{def}{=} \sum_x p(x) \log \frac{p(x)}{q(x)} \quad (3)$$

The divergence satisfies three properties which are referred as the divergence properties:

(a) The K-L divergence is zero when self similarity will be occurred.

(b) The K-L divergence is zero when two probability density functions would be identical.

(c) The K-L divergence is always non-negative for all *p, q*.

For two Gaussians $\hat{p}$ and $\hat{q}$, the K-L divergence has a closed formed expression,

$$KL(\hat{p} \| \hat{q}) = \frac{1}{2}\left[\log\frac{|\Sigma_{\hat{q}}|}{|\Sigma_{\hat{p}}|} + Tr[\Sigma_{\hat{q}}^{-1} \Sigma_{\hat{p}}] - d + (\mu_{\hat{p}} - \mu_{\hat{q}})^T \Sigma_{\hat{q}}^{-1} (\mu_{\hat{p}} - \mu_{\hat{q}})\right] \qquad (4)$$

whereas for two GMMs, no such closed form expression exists. Here, *p* and *q* to be GMMs. The marginal densities of *x* ε $R^d$ under *p* and *q* are

$$p(x) = \sum_a \pi_a N(x; \mu_a; \Sigma_a)$$
$$q(x) = \sum_b \omega_b N(x; \mu_b; \Sigma_b) \qquad (5)$$

where $\pi_a$ is the prior probability of each state, $N(x; \mu_a; \Sigma_a)$ is a Gaussian in *x* with mean $\mu_a$ and variance $\Sigma_a$.

Due to the large complexity it is not possible to compute the K-L divergence directly. Therefore the Gaussian approximations can be defined as follows

$$KL(p(x) \| q(x)) \cong \sum_{i=1}^{N} P_i \min_j (KL(f_p(x|i) \| f_q(x|j)) + \log\left[\frac{p_i(x)}{q_j(x)}\right] \qquad (6)$$

where, $f_p(x|i)$ and $f_q(x|i)$ are multivariate Gaussians in Equation (6).

After modeling the ear image using Gaussian mixture model and finding similarity between the ears images using K-L divergence, the pixels are now classified into several clusters or regions of identical approximated color features. The algorithm discussed in this section partitions the color pixels within the mask area or within the cropped region of ear image. GMM is used to fit the pixels of unmasked regions in the image. Each slice region is represented by multivariate Gaussian and the weighted collection of multivariate Gaussians approximate the distribution of pixels of cropped ear images.

In this work, an assumption has been made that each ear image can produce different number of slice regions. However, the number of extracted slice regions may have identical with their approximated locations for multiple ear instances of a subject.

## 4. SIFT KEYPOINTS EXTRACTION

The Scale Invariant Feature Transform (SIFT) descriptor [12] is invariant to image rotation, scaling, partly illumination changes and the 3D camera view. The investigation of SIFT features for biometric authentication has been explored further in [5,7,17]. Initially, a pyramid of images is created by convolving the original image by a set of Difference-of-Gaussian (DOG) kernels. The difference of the Gaussian function is increased by a factor in each step. The SIFT descriptor detects feature points efficiently through a staged filtering approach that identifies stable points in the scale-space of the resulting image pyramid. Local feature points are extracted through selecting the candidates for feature points by searching peaks in the scale-space from a DoG function. Further the feature points are localized using the measurement of their stability and assign orientations based on local image properties. Finally, the feature descriptors, which represent local shape distortions and illumination changes, are determined.

Prior to feature extraction from color slice regions, slice regions are converted into grayscale slice regions by using the technique presented in [23]. The model ear image is normalized by histogram equalization and then SIFT features [12] are extracted from the color slice regions. Each feature point contains four types of information – spatial location (*x, y*), scale (*S*), orientation (*θ*) and Keypoint descriptor (*K*). For the experiment, only keypoint descriptor [12] information has been used which consists of a vector of 128 elements representing neighborhood intensity changes of each keypoint. More formally, local image gradients are measured at the selected scale in the region around each keypoint. The measured gradients information is then transformed into a vector representation that contains a vector of 128 elements for each keypoints calculated over extracted keypoints. These keypoint descriptor vectors represent local shape distortions and changes due to illumination.

## 5. FUSION STRATEGY OF KEYPOINT FEATURES

In the proposed ear recognition model, detected SIFT features from color-segmented slice regions are fused together by concatenation and Dempster-Shafer decision theory. The keypoints are extracted from different slice regions are taken to make an augmented group of features for both the reference ear model and the probe ear model. The proposed fusion strategies use feature level fusion approaches which are used to fuse the feature sets obtained from different color segmented slice regions.

### 5.1 Keypoints fusion using concatenation approach

Assume, a segmented ear image $I$ is given where the independent color slice regions $S$ are extracted by the method discussed in Section 3. In each slice region the SIFT feature points are varying. After extraction of the SIFT feature points from the segmented slice regions the feature points are gathered together by concatenation [21] into an augmented group for each reference model and probe model. Finally, the matching between these two sets of augmented groups is performed using Euclidean distance approach. While matching are accomplished between a pair of segmented ear images, the matching scores are obtained and decision for user verification is done by comparing the score against a threshold ($\psi$). In order to obtain fused sets of features for both the reference and the probe models, the keypoints are detected in varying number for each segment region as $K_1, K_2, K_3, \ldots K_S$. Now, an augmented set is obtained $DS$ of SIFT features by concatenation as follows

$$DS = \{K_1 \cup K_2 \cup K_3 \cup \ldots \cup K_S\} \tag{7}$$

The feature set $DS$ represents the proximity among detected SIFT features of the color slice regions.

Finally, the final matching distance $D_{final}$ ($DS_{probe}$, $DS_{reference}$) is computed on the basis of the number of keypoints paired between two sets of features and is given by

$$D_{final} = \sqrt{\sum_{i \in D_{probe}, j \in D_{reference}} (DS_{probe}(K_i) - DS_{reference}(K_j))^2} \leq \Psi \tag{8}$$

where $DS_{probe}$ and $DS_{reference}$ are the concatenated feature sets for both the probe model and the reference model and $\psi$ is the threshold determined from a subset of database. As for the matching threshold, this ear set is disjoint from the image sets used for testing and validation.

### 5.2 Keypoints fusion using Dempster-Shafer decision theory

The Dempster-Shafer decision theory [7] is used to integrate the detected keypoint features obtained from individual slice regions. It is based on combining the evidences obtained from different sources to compute the probability of an event. This is obtained by combining three elements: the basic probability assignment function (*bpa*), the belief function (*bf*) and the plausibility function (*pf*).

The *bpa* maps the power set to the interval [0,1]. The *bpa* function of the empty set is 0 while the *bpa's* of all the subsets of the power set is 1. Let $m$ denote the *bpa* function and $m(A)$ represent the *bpa* for a particular set $A$. Formally, the basic probability assignment function can be represented by the following equations

$$m : \breve{A} \rightarrow [0,1] \tag{9}$$

$$m(\emptyset) = 0 \tag{10}$$

$$\sum_{A \in \breve{A}} m(A) = 1 \tag{11}$$

where $\breve{A}$ is the power set of A and $\emptyset$ is the empty set. From the basic probability assignment, the upper and lower bounds of an interval are bounded by two non-additive continuous measures called, Belief and Plausibility. The lower bound, Belief, for a set A is defined as the sum of all the basic probability assignments of the proper subsets B of the set of interest A. The upper bound, Plausibility, is the sum of all the basic probability assignments of the sets B that intersect A. Thus, Belief for a A, Bel(A) and Plausibility of A, Pl(A) can be defined as

$$Bel(A) = \sum_{B|B \subseteq A} m(B) \qquad (12)$$

$$Pl(A) = \sum_{B|B \cap A \neq \varnothing} m(B) \qquad (13)$$

An inverse function with the Belief measures can be used to obtain the basic probability assignment. Therefore,

$$m(A) = \sum_{B|B \subseteq A} (-1)^\gamma Bel(B) \qquad \because \gamma = |A - B| \qquad (14)$$

where |A-B| is the difference of the cardinality between the two sets A and B. It is possible to derive these two measures, Belief and Plausibility from each other with the help of following equation

$$Pl(A) = 1 - Bel(\overline{A}) \qquad (15)$$

where $\overline{A}$ is the complement of A. In addition, the Belief and Plausibility measures can be written as:

$$Bel(\overline{A}) = \sum_{B|B \subseteq A} m(B) = \sum_{B|B \cap A = \varnothing} m(B) \qquad (16)$$

and

$$\sum_{B|B \cap A \neq \varnothing} m(B) = 1 - \sum_{B|B \cap A = \varnothing} m(B) = Pl(A) \qquad (17)$$

Let $\Gamma^{SR1}, \Gamma^{SR2}, \Gamma^{SR3}, \ldots, \Gamma^{SRn}$ be the feature sets obtained from the *n* numbers of slice regions. The dimensions of feature points in all sets of slice regions may be different. To make the features sets of equal dimensions, the features set with less numbers of features points added with zero to the remaining points and make it equal length with other feature sets. Now, in order to obtain the transformed combine feature set Dempster combination rule is applied. Also, let $m(\Gamma^{SR1})$, $m(\Gamma^{SR2}), m(\Gamma^{SR3}), \ldots, m(\Gamma^{SRn})$ be the *bpa* functions for the Belief measures $Bel(\Gamma^{SR1}), Bel(\Gamma^{SR2}), Bel(\Gamma^{SR3}), \ldots, Bel(\Gamma^{SRn})$ for the *n* numbers of slice regions respectively. Then the Belief probability assignments (*bpa*) can be combined together to obtained a Belief committed to a feature set $C \in \Theta$ according to the following combination rule or orthogonal sum rule

$$m(C_1) = m(\Gamma^{SR1}) \oplus m(\Gamma^{SR2}) = \frac{\sum_{\Gamma^{SR1} \cap \Gamma^{SR2} = C} m(\Gamma^{SR1}) m(\Gamma^{SR2})}{1 - \sum_{\Gamma^{SR1} \cap \Gamma^{SR2} \neq \varnothing} m(\Gamma^{SR1}) m(\Gamma^{SR2})}, \quad C_1 \neq \varnothing. \qquad (18)$$

$$m(C_2) = m(\Gamma^{SR3}) \oplus m(\Gamma^{SR4}) = \frac{\sum_{\Gamma^{SR3} \cap \Gamma^{SR4} = C_2} m(\Gamma^{SR3}) m(\Gamma^{SR4})}{1 - \sum_{\Gamma^{SR3} \cap \Gamma^{SR4} \neq \varnothing} m(\Gamma^{SR3}) m(\Gamma^{SR4})}, \quad C_2 \neq \varnothing. \qquad (19)$$

……….  …………..  ………….  ……………  …………..
……….  …………..  ………….  ………….  …………..

$$m(C_n) = m(\Gamma^{SR(n-1)}) \oplus m(\Gamma^{SRn}) = \frac{\sum_{\Gamma^{SR(n-1)} \cap \Gamma^{SRn} = C_n} m(\Gamma^{SR(n-1)}) m(\Gamma^{SRn})}{1 - \sum_{\Gamma^{SR(n-1)} \cap \Gamma^{SRn} \neq \varnothing} m(\Gamma^{SR(n-1)}) m(\Gamma^{SRn})}, \quad C_n \neq \varnothing. \qquad (20)$$

Let, $m(C_1), m(C_2), \ldots, m(C_n)$ be obtained from different sets of pairs of features. They can be further fused by the following equation

$$S = m(C_1) \oplus m(C_2) \oplus m(C_3) \oplus .... \oplus m(C_n) \qquad (21)$$

The denominators of Equations from (18) to (20) denote the normalizing factors which denote the art of belief assignments. The notation $\oplus$ denotes the Dempster combination rule. The combined final value $S$ represents the transformed feature value and representative feature of all keypoint features.

Let, $S_1$ and $S_2$ be the final transformed feature values obtained from reference ear model and probe ear model respectively. Finally, the final matching distance $D'_{final'}(S_1, S_2)$ is computed on the basis of the number of keypoints paired between two sets of features. The similarity score can be computed as follows

$$D'_{final} = \sqrt{\sum (S_2 - S_1)^2} \qquad (22)$$

The final decision of user acceptance and rejection can be established by applying the threshold $\Phi$ to the final matching score $D'_{final'}$.

$$decision = \begin{cases} accept, & if \; D'_{final'} \leq \Phi \\ reject, & otherwise \end{cases} \qquad (23)$$

where $\Phi$ is a predefined threshold value.

## 6. EXPERIMENTAL RESULTS

### 6.1 Identification Performance

The proposed identification technique has been tested on IIT Kanpur ear database [17]. The database consists of 800 ear images of 400 individuals and all the frontal view ear images are considered for evaluation. The ear images are taken under controlled environment in different sessions. The ear viewpoints are consistently kept neutral and the ear images are downscaled to 200×240 pixels with 500 dpi resolution. High resolution ear image helps to increase the SIFT keypoints during feature extraction. Ear images are acquired by high resolution digital camera. The entire database is divided into two different groups such as reference and probe. For reference model single ear image is enrolled for each individual. Therefore, 400 ear images are considered for training session. The remaining 400 images are used for testing and evaluation. Sample ear images are shown in Figure 1.

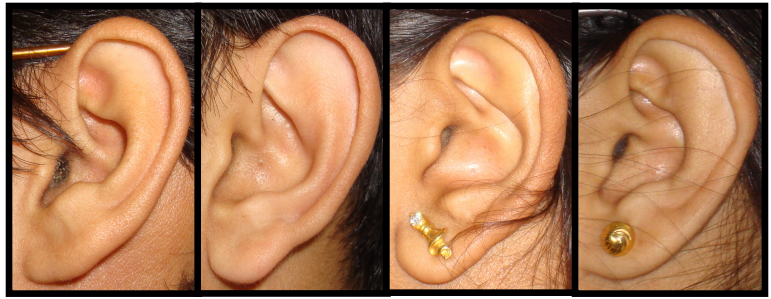

Figure 1. Sample Ear Images from IIT Kanpur Database

The experiments are conducted in two sessions. In the first session, the ear identification is performed with SIFT features before color segmentation into slice regions and in the next session, identification is performed with the detected SIFT keypoint features from segmented slice regions. During identification, each probe ear image is matched against all the ear images in the reference set and matching probabilities are obtained for all probe images. In this identification process, top best 5 matching probabilities are considered while it is also ensure that the matching probability for probe ear is lied within these top 5 closest matched according to descending order of matching proximities. From these matching proximities Cumulative Match Characteristics (CMC) curve is drawn which a trade-off between identification probability and rank of the probe set. Rank order statistics and cumulative match characteristics (CMC) curve are useful to measure the performance of the proposed identification system. The matching probability of the database represents the rank obtained while matching is done with a probe ear.

Table 1. Identification Rates for the Proposed Ear Identification Systems

| METHODS ↓ PERFORMANCE → | IDENTIFICATION RATE (%) |
|---|---|
| Prior to color segmentation | 92.5 |
| After color segmentation (Concatenation fusion rule) | 94.75 |
| After color segmentation (DS theory based fusion rule) | 98.25 |

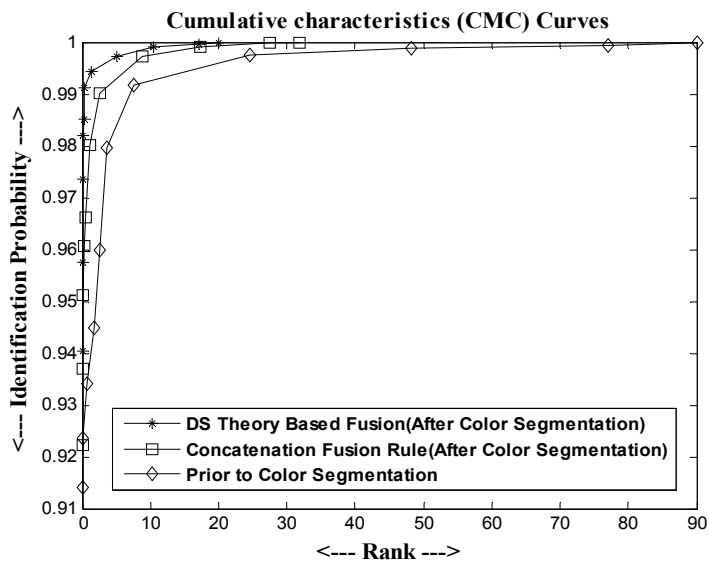

Figure 2. Cumulative Match Characteristics Curves (CMC) for the Proposed Methods

The identification accuracy is determined by computing the matching proximities of the probe samples and set these values against the ranks obtained from CMC curve. The identification rate for the concatenation fusion rule found to be 94.75% while that based on Dempster-Shafer decision theory is 98.25%. The accuracies are obtained after color segmentation of ear image while prior to color segmentation of ear image identification accuracy is 92.5%. Color segmentation is found to be one of the efficient approaches to increase the overall identification performance while in most of the cases, other ear recognition techniques found to be inconsistent with the increase in samples in the database.

More false matches have been found from the non-segmentation ear skin than the color segmentation approach and due to segmentation of ear image with high resolution image, more number of SIFT keypoints can be found. The performance of the proposed system is shown in the Table 1 and the corresponding accuracy curves are drawn in the CMC curve in Figure 2. The effect of SIFT feature descriptor is also found to be robust to the proposed approach.

### 6.2 Verification Performance

It has been determined that when Euclidean distance metric is used for verification before color segmentation, the system achieved accuracy of 91.09% with False Positive (FP) and True Negative (TN) of about 9.56% and 8.26% respectively and when nearest neighbor approach is used for verification before color segmentation, the system achieves 93.01% recognition accuracy with FP and TN of about 4.38% and 9.6%, respectively. Due to some false matches from non-segmented slice regions, recognition accuracy often degrades. On the other hand, when the ear image is segmented into some color slice regions, overall system accuracy is increasing radically. False pair SIFT keypoints are often found in non-segmented color slice regions. So, to minimize the false match pairs of SIFT keypoints corresponding to a pair of ear images and to increase the true match pairs of keypoints, ear image is segmented into a number of segmented slice regions. These color segmented slice regions represent that area where most of the true match keypoints are found. When Euclidean distance is used for verification with these segmented slice regions only, the ear recognition system achieved 94.31% recognition accuracy with FP is of 4.22% and TN is of 7.16% respectively. In another case, when nearest neighbor is used for verification with these slice regions, the system achieves 96.93% recognition accuracy and the computed FP and TN error rates are 2.14% and 4%, respectively. For Dempster-Shafer theory based fusion rule is used to fuse the detected keypoints from the slice regions with Euclidean distance measure, 97.5% recognition rate is recorded, while FP and TN are determined as 1.75% and 3.25% respectively. Table 2 shows the ear verification performance for the proposed system using SIFT features and color segmentation methodology. The ROC curves for different experiments are shown in Fig. 3.

Table 2. Verification, FP, TN Rates for the Different Methods

| METHOD | DISTANCE MEASURE | VERIFICATION ACCURACY (%) | FALSE POSITIVE (%) | TRUE POSITIVE (%) |
|---|---|---|---|---|
| Prior to color segmentation | Euclidean Distance | 91.09 | 9.56 | 8.26 |
| | Nearest Neighbor | 93.01 | 4.38 | 9.60 |
| After color segmentation (Concatenation rule) | Euclidean Distance | 94.31 | 4.22 | 7.16 |
| | Nearest Neighbor | 96.93 | 2.14 | 4.00 |
| After color segmentation (DS rule) | Euclidean Distance | 97.5 | 1.75 | 3.25 |

Apart from the robust performance of the proposed system, the results have been compared with the ear recognition systems that are developed using the SIFT features. In [5], a SIFT-based ear recognition work has been proposed which achieves 96% recognition accuracy for baseline recognition. In addition, another set of recognition results have been presented by considering some variations in nearby clutter, occlusion and pose changes to ear images. However, the accuracy achieved in this work gives 96.93%, which outperforms the ear recognition technique discussed in [5,22]. In [16], the authors have developed an ear recognition system, which uses SIFT descriptor for feature extraction from multi-pose ear images and finally, recognition accomplishes by template fusion. However this method also outperforms the work presented in [16]. Table 3 shows a comparative study where the proposed work outperforms the works presented in [5,16,22].

Color segmentation of ear images into several color similarity slice regions not only reduce the false matches by discarding non-segmented color regions but also achieves desired recognition accuracy.

Table 3. Comparison of the Proposed Verification Methods with the Methods Presented in [5,16,22]

| METHODS | RECOGNITION RATES |
|---|---|
| Ear Recognition – Feature Fusion Method [16] | 95.13% |
| Ear Registration and Recognition Method [5] | 96% |
| Model based Ear Recognition [22] | 91.5% |
| Color Segmentation based Ear Recognition [Concatenation - NN Distance Approach] | 96.93% |
| Color Segmentation based Ear Recognition [DS Fusion] | 97.5% |

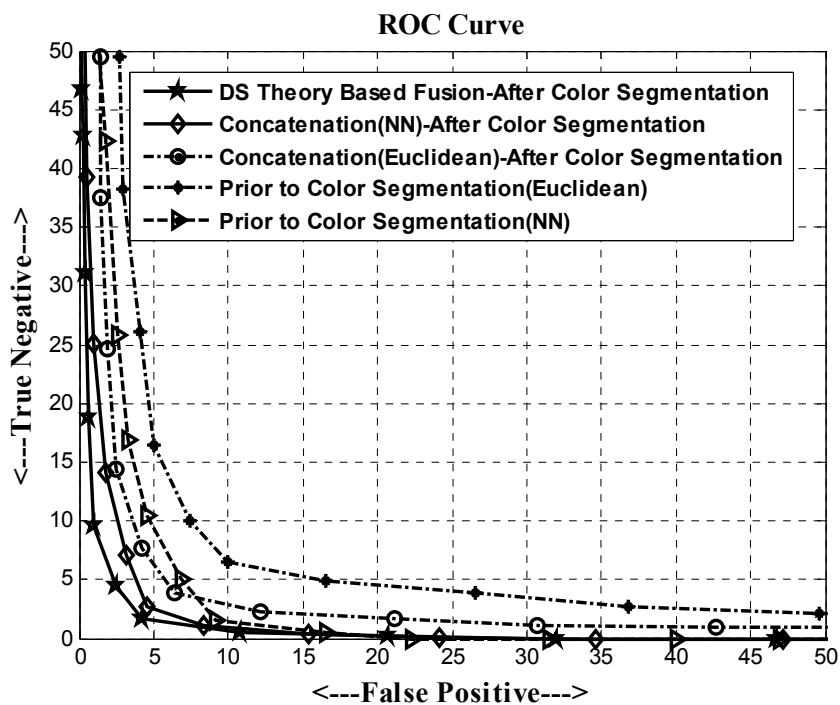

Figure 3. Receiver Operating Characteristics (ROC) Curves for Different Proposed Methods

## 7. CONCLUSION

This paper has presented efficient ear identification and verification system which uses SIFT descriptor for feature extraction from color similarity slice regions. The color segmented regions are considered as the area from which the maximum number of keypoints is found as matching points. The remaining parts of the ear image which are not included in color slice regions are discarded. As a result the proposed system is found to be robust also the Gaussian mixture model framework with K-L divergence proves to be a better characterization for successfully dividing the ear image into a number of segmented regions. These segmented regions acts as the high probability matching regions for SIFT features. The experiments have been conducted in two different sessions. In the first session, results are obtained prior to segmentation of ear images and are supported through the two distance metrics. In the next session, segmented regions are considered for feature extraction and matching SIFT keypoint features. It has used two different fusion rule namely, concatenation and Dempster combination rule for fusing the keypoint features extracted from slice regions. During identification and verification in the second session, the system achieves with an accuracy of more than 97%. This proves that the system can be deployed to high security applications where single modality can be used with efficient and cost effective manner.